\documentclass[journal]{IEEEtran}
\usepackage{amsmath,amsfonts}
\usepackage{algorithmic}
\usepackage{algorithm}
\usepackage{array}
\usepackage[caption=false,font=normalsize,labelfont=sf,textfont=sf]{subfig}
\usepackage{textcomp}
\usepackage{stfloats}
\usepackage{url}
\usepackage{verbatim}
\usepackage{graphicx}
\usepackage{cite}
\begin{document}

\title{SITSMamba for Crop Classification based on Satellite Image Time Series }

\author{Xiaolei Qin,~\IEEEmembership{Graduate Student Member,~IEEE,}
Xin Su,~\IEEEmembership{Member,~IEEE,}
Liangpei Zhang,~\IEEEmembership{Fellow,~IEEE}
        
\thanks{Xiaolei Qin, and Liangpei Zhang are with the State Key Laboratory of Information Engineering in Surveying, Mapping and Remote Sensing,
 Wuhan University, Wuhan 430079, China (e-mail: qinxlei@whu.edu.cn;
 zlp62@whu.edu.cn).
 \par Xin Su is with the School of Remote Sensing and Information Engineering,
 Wuhan University, Wuhan 430079, China (e-mail: xinsu.rs@whu.edu.cn).
 }
}

\markboth{Journal of \LaTeX\ Class Files,~Vol.~14, No.~8, August~2021}%
{Shell \MakeLowercase{\textit{et al.}}: A Sample Article Using IEEEtran.cls for IEEE Journals}


\maketitle

\begin{abstract}
Satellite image time series (SITS) data provides continuous observations over time, allowing for the tracking of vegetation changes and growth patterns throughout the seasons and years. Numerous deep learning (DL) approaches using SITS for crop classification have emerged recently, with the latest approaches adopting Transformer for SITS classification. However, the quadratic complexity of self-attention in Transformer poses challenges for classifying long time series. While the cutting-edge Mamba architecture has demonstrated strength in various domains, including remote sensing image interpretation, its capacity to learn temporal representations in SITS data remains unexplored. Moreover, the existing SITS classification methods often depend solely on crop labels as supervision signals, which do not sufficiently emphasize the importance of temporal information for the network. In this paper, we proposed a Satellite Image Time Series Mamba (SITSMamba) method for crop classification based on remote sensing time series data. The proposed SITSMamba contains a spatial encoder based on Convolutional Neural Networks (CNN) and a Mamba-based temporal encoder. To exploit richer temporal information from SITS, we design two branches of decoder used for different tasks. The first branch is a crop Classification Branch (CBranch), which includes a ConvBlock to decode the feature to a crop map. The second branch is a SITS Reconstruction Branch that uses a Linear layer to transform the encoded feature to predict the original input values. Furthermore, we design a Positional Weight (PW) applied to the RBranch to help the model learn rich latent knowledge from SITS. We also design two weighting factors to control the balance of the two branches during training. When evaluated on the PASTIS32 and MTLCC datasets, our SITSMamba achieves an OA of 0.7416 and 0.9104, respectively, outperforming the previous state-of-the-art (SOTA) methods. The code of SITSMamba is available at: https://github.com/XiaoleiQinn/SITSMamba.
\end{abstract}


\section{Introduction}
\IEEEPARstart{C}{rop} monitoring with satellite images has become a hot topic in the field of remote sensing\cite{lin2022early,wolanin2019estimating}. Identifying and mapping the spatial distribution of crop types enables efficient agricultural management in a large scale. This process plays a critical role in optimizing agricultural productivity, resource management, and decision-making\cite{yuan2023bridging,cole2018science}.

Time series information is critical for crop type identification, since crop types usually exhibit distinct phenological patterns \cite{xu2021towards}. Due to the rapid development of remote sensing, it is now easy to access to satellite image time series (SITS) with high temporal resolution. The repeated observations from remote sensing instruments allows dynamical monitoring of crop growth and provides important temporal information for crop type classification. 

To utilize the temporal information for crop type classification, a series of traditional methods including threshold-based approaches and pre-defined mathematical equations have been proposed \cite{fan2014characterizing,foerster2012crop,sakamoto2010two}. Moreover, machine learning models such as Support Vector Machine (SVM), and Random Forest (RF), have also been adopted for identifying crop types \cite{pelletier2016assessing}. Although these machine learning models can extract phenology information of crops, their performances heavily rely on manual feature selection and are difficult to apply to other regions.

The emergence of deep learning (DL) has introduced innovative models for SITS classification, significantly enhancing the efficiency and accuracy of crop mapping. To enhance the modeling ability of the temporal context, these DL methods process temporal information mainly based on three types of model structures: Recurrent Neural Networks (RNNs), Convolutional Neural Networks (CNNs), and Transformers \cite{vaswani2017attention}. RNNs, known for their recursive connections, excel at capturing temporal correlations and allowing information to flow across sequential cells, which has led to their widespread application in crop classification using time series data\cite{ienco2017land,yuan2020using,russwurm2017temporal}. CNN-based methods offer a different yet effective strategy for time series data analysis by stacking convolutional layers to detect temporal patterns within the input series \cite{zhong2019deep}. In contrast, the Transformer \cite{vaswani2017attention} architecture processes all temporal positions in parallel by calculating attention scores to model the interactions between positions. With its great ability to extract long-term dependencies, more and more methods adopt Transformer for crop mapping tasks\cite{russwurm2020self,garnot2021panoptic,yuan2020self}. However, Transformer's self-attention mechanism has the issue of quadratic complexity, which results in a high computational burden and impedes the extraction of long-term features. 

State-space models (SSMs) \cite{smith2022simplified,ma2022mega} have recently become popular due to their capability to efficiently handle continuous long-sequence data analysis, which is a common challenge in a wide range of fields, including natural language processing and computer vision. As a typical type of SSM, the S4 model has been further advanced with the introduction of Mamba \cite{gu2023mamba}, which incorporates a selective mechanism allowing the model to select relevant information based on the input. This design enhances its computational efficiency compared to the widely used Transformer model when dealing with long sequence data. Mamba's efficiency of processing long sequences makes it a promising candidate for applications in SITS analysis. However, to the best of our knowledge, no study has yet explored the application of Mamba in the classification of SITS data.

\begin{figure}
	\flushleft 
	\includegraphics[scale=0.85]{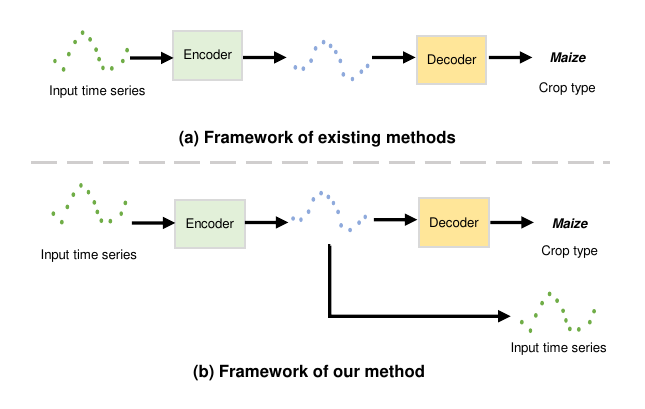}
	\caption{Overview of SITSMamba.}
	\label{introframework}
\end{figure}

Although deep learning models are continuously improving and achieving higher accuracy in remote sensing image time series classification tasks, there are still instances of misclassification across categories. This challenge often stems from the similar phenological patterns across various crop types \cite{russwurm2018multi}. Moreover, the complex temporal pattern of the crop makes it difficult for models to extract robust temporal features, which can easily lead to model overfitting and poor generalization. In the previous fully-supervised methods, only category labels are used to supervise the spatiotemporal feature learning (as shown in Fig.~\ref{introframework}(a). However, it may be difficult to map a long and complex time series to a category. \textbf{A natural question arises: apart from crop label, is there any other information can be used for supervising the model to learn robust representation from the complex temporal patterns?}    

To tackle the aforementioned problems, we design a SITS learning model that combines CNN and Mamba as spatial encoder and temporal encoder, respectively. To the best of our knowledge, this is the first time that Mamba is used for SITS classification. Furthermore, we suppose that apart from crop type label, the SITS itself can serve as a supervision signal. This hypothesis is partly supported by the recent self-supervised learning (SSL) methods applied to SITS, which involve performing a masking and reconstructing task on the original data \cite{yuan2020self,yuan2022sits,dumeur2024self}. Different from the two-step frameworks typical in SSL, we integrate the SITS reconstruction task directly into the crop classification workflow. Consequently, we design a multi-task SITS learning framework (as shown in Fig.~\ref{introframework}(b)). Our method includes a crop Classification Branch (CBranch) and SITS Reconstruction Branch (RBranch). The supervision of RBranch can enhance the learning of the spatial and temporal encoder during training, which further improves the training of CBranch. Assuming that prediction difficulty decreases with later temporal positions, we design a Positional Weight (PW) applied to the loss of RBranch.  The contributions of this article are summarized as follows.
\begin{enumerate}
    \item We developed a SITSMamba that adopts Mamba as temporal encoder for crop classification based on SITS. Combined with CNN as the spatial encoder, SITSMamba outperforms previous methods in crop classification.

    \item We built a multi-task learning framework for SITS classification, designed to enhance the performance of crop mapping. This framework contains two branches of decoder, a CBranch and an RBranch. The CBranch decodes the features into crop type probability, while the RBranch reconstructs the original SITS. The auxiliary learning of RBranch enhances the spatiotemporal encoding, thereby subsequently improving the accuracy of crop classification.

    \item To achieve better supervision of RBranch, we designed a Positional Weight (PW) that increases the weight of loss as the temporal position increases. Additionally, to balance the learning of CBranch and RBranch, we designed two weights applied to the loss of RBranch.
\end{enumerate}

\section{Related Work}\label{sec:2}

\subsection{Crop classification with SITS}

\subsubsection{Traditional and ML methods}The traditional crop classification approaches adopt threshold, or pre-defined functions to classify crop patterns. Foerster \emph{et al.} \cite {foerster2012crop} produce temporal profiles of crop types and define a rule to classify them according to the NDVI values. Fan \emph{et al.} \cite{fan2014characterizing} introduced an adaptive threshold technique to identify crop intensity. Sakamoto \emph{et al.} \cite {sakamoto2010two} utilized wavelet-based smoothing and shape-model fitting on time-series data to pinpoint phenological stages of maize and soybean. Relying on the observation that winter crops have established biomass by early spring, Skakun \emph{et al.} \cite{skakun2017early} developed a Gaussian mixture model (GMM) to discriminate the season of crops. These methods need relevant agricultural knowledge to identify certain crop types. Some other machine learning (ML) methods, such as Random Forest (RF), and Support Vector Machine (SVM), have been explored to achieve crop classification more automatically. Zhao \emph{et al.} \cite{zhao2020robust} developed a spectral-spatial agricultural crop mapping method using conditional random fields (SCRF) that leverages spectral information and spatial pixel interactions. Li \emph{et al.} \cite{li2023development} employed a random forests approach combined with a multi-scale, multi-temporal mapping procedure to map maize and soybean.

\subsubsection{DL methods}Recent research has focused on developing DL models for crop classification to eliminate the need for laborious manual feature selection processes. Pelletier \emph{et al.}\cite{pelletier2019temporal} proposed Temporal Convolutional Neural Networks (TempCNNs) that employ temporal convolutions to autonomously extract features from temporal data. Based on RNN networks, Ienco \emph{et al.} \cite{ienco2017land} explored the ability of long short-term memory (LSTM) for both pixel-based and object-based time series classifications. Considering both spatial and temporal information, Ru{\ss}wurm \emph{et al.} \cite{russwurm2018multi} developed a convolutional LSTM network for SITS classification. Similarly, Garnot \emph{et al.} \cite{garnot2019time} designed a hybrid recurrent convolutional network to learn spatial and temporal features simultaneously.  Recently, Transformer \cite{vaswani2017attention} has been widely used in sequential or image data because of their abilities to model long-range dependencies and global spatial features. As the key component of Transformer, self-attention (SA) mechanism was introduced to learn temporal patterns for crop classification \cite{garnot2020satellite, russwurm2020self}. Moreover, the SA temporal feature extractor was also combined with a spatial CNN-based encoder to achieve spatiotemporal feature learning \cite{garnot2021panoptic}. Recognizing that Transformers can be applied not only to the temporal dimension but also to learn spatial context, Tarasiou \emph{et al.} \cite{tarasiou2023vits} developed a pure Transformer architecture. This method used ViT architecture to model the spatial context and adopted Transformer to learn temporal information. 

To further enhance the spatiotemporal representation, a variety of SSL methods have been developed for SITS. Yuan and Lin \cite{yuan2020self} introduced a pioneering self-supervised pre-training strategy utilizing pixel-level temporal masking, addressing the challenge of scarce labeled samples in crop classification tasks. To consider the spatial information, Yuan \emph{et al.}\cite{yuan2022sits} presented SITS-Former, a model pre-trained via SSL for patch-based representation learning and classification of SITS data. More recently, Dumeur \emph{et al.}\cite{dumeur2024self} proposed U-BARN, a fully convolutional architecture designed to learn spatio-spectral image representations and employing Transformers to harness temporal relationships. These SSL methods demonstrate that SITS data can serve as a supervision signal, aiding the model in learning valuable spatiotemporal representations. Unlike these SSL methods, which utilize SITS masking and reconstruction as a pretext task, we have integrated SITS reconstruction directly into the crop classification workflow and developed a multi-task framework.

\subsection{State-Space Models (SSMs)}
\subsubsection{SSMs for Sequence}
The Structured State-Space Sequence (S4) model has demonstrated its capability to capture long-range dependencies which is particularly important in processing sequential data. Whereas an S4 layer uses independent single-input, single-output SSMs, Smith \emph{et al.} \cite{smith2022simplified} enhanced its capabilities by integrating Multi-Input Multi-Output State-Space Models and an efficient parallel scan operation, resulting in the creation of the S5 layer. This advancement aimed to boost the model's performance while maintaining computational efficiency. In the realm of language modeling, Mehta \emph{et al.} \cite{mehta2022long} introduced the Gated State Space (GSS) layer for sequence modeling for English books, GitHub source code, and ArXiv mathematics articles. This method showed competitive performance against Transformer baselines and improved results through the integration of self-attention for local dependency modeling. The S4 model also shows potential for modeling complex spatiotemporal dependencies in videos. However, its equal treatment of all image tokens can limit efficiency and accuracy. To overcome this, a Selective S4 (S5) model is introduced with a mask generator for selective token processing to enhance the model's performance in video understanding tasks \cite{wang2023selective}. Recently, \cite{gu2023mamba} proposed Mamba, a novel neural network architecture that enhances sequence modeling by integrating selective structured state space models (SSMs). This mechanism allows for content-based reasoning and efficient processing of long sequences, outperforming Transformers of the same size or even larger. 
\subsubsection{Application of Mamba in Remote Sensing}
Recently, the Mamba architecture has been successfully integrated into remote sensing applications. Specifically, Chen \emph{et al.} \cite{chen2024changemamba} adapted Mamba for remote sensing change detection tasks through three specialized frameworks. These frameworks surpass current CNN and Transformer approaches on benchmark datasets, showcasing the Mamba architecture's effectiveness in capturing spatio-temporal relationships for various change detection scenarios. Zhao \emph{et al.} \cite{zhao2024rs} proposed a Remote Sensing Mamba (RSM) model for dense prediction tasks in large very-high-resolution (VHR) remote sensing images, featuring an omnidirectional selective scan module (OSSM) for capturing global context. Apart from VHR data, Mamba structure has also been applied to hyperspectral image (HSI) \cite{he20243dss,yao2024spectralmamba}. For example, \cite{li2024mambahsi} proposed a MambaHSI model to enhance hyperspectral image classification by designing specialized spatial and spectral Mamba blocks for integrated feature extraction. This approach demonstrates superior performance and efficiency in HSI classification tasks across various datasets. As for low-level tasks, \cite{fu2024ssumamba} proposed SSUMamba, a novel denoising model for hyperspectral images that utilizes the spatial-spectral continuous scan (SSCS) Mamba architecture to effectively model long-range dependencies and correlations. 

Motivated by the effective integration of Mamba in VHR and hyperspectral remote sensing imagery, we investigate the application of Mamba blocks for capturing long time series data in remote sensing. Additionally, we introduce a multi-branch decoder framework designed to convert Mamba's output into either crop types or reconstructed SITS data.

\section{Methodology}\label{sec:3}
\begin{figure*}
	\centering
	\includegraphics[scale=0.7]{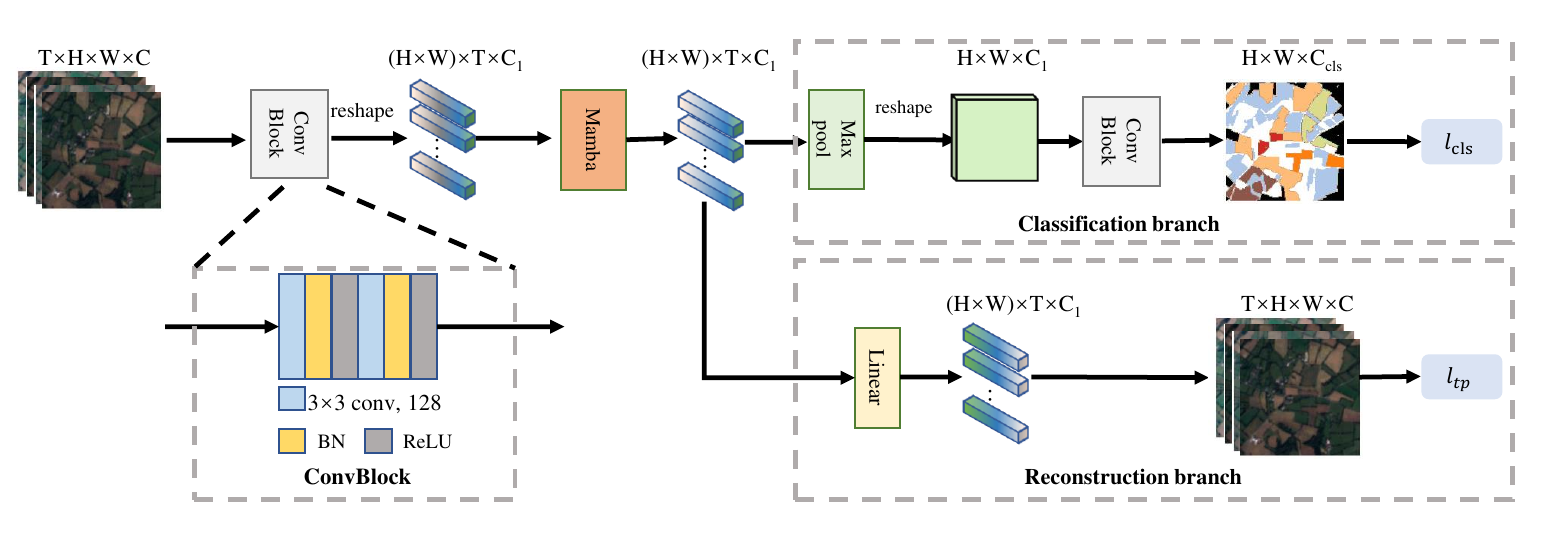}
	\caption{Overview of SITSMamba.}
	\label{architecture}
\end{figure*}
\subsection{Overview Architecture}
An overview of the proposed SITSMamba is shown in Fig.~\ref{architecture}. The framework contains three main components: embedding layer, encoder backbone, and segmentation head. The proposed SITSMamba adopts CNN as the spatial encoder and Mamba block as the temporal encoder. The spatial encoder is a Convolutional block (ConvBlock), which is used in the UTAE model \cite{garnot2021panoptic}. This ConvBlock extracts spatial feature simultaneously from multi-temporal images. It contains a series of convolutional layers, Rectified Linear Unit (ReLU) activations, and normalization layers. Then the SITS data is reshaped and input to the Mamba block. The framework includes two branches, a Classification Branch (CBranch) and a Reconstruction Branch (RBranch). The CBranch uses a max pooling on temporal dimension and generates a mono-temporal image after reshaping. This step aims to produce a feature map with rich information related to crop types. To enhance the spatial and temporal encoding, we design an RBranch which aims to transform the encoded feature map to the original SITS. Both the CBranch and the RBranch are used in the training process, whereas during inference, only the CBranch is used for crop classification and the RBranch is ignored.

\subsection{SSM mechanism}
As the key mechanism of Mamba, SSM-based models represent dynamic systems that transform an input sequence $x\left(t\right) \in \mathcal{R}$ to a response $y\left(t\right) \in \mathcal{R}$ through a hidden state $h\left(t\right) \in \mathcal{R}^{N}$. These systems can be formulated as linear ordinary differential equations (ODEs) 
\begin{equation}
\left\{
    \begin{aligned}
      &h^{\prime}\left(t\right) =\mathbf{A} h(t)+\mathbf{B} x(t)  \\
      &y(t) =\mathbf{C} h(t)
    \end{aligned}
\right.
\label{eq:ODEs}
\end{equation}
where $\mathrm{\mathbf{A}}\in \mathcal{R}^{N \times N}$, $\mathrm{\mathbf{B}} \in \mathcal{R}^{N\times 1}, \mathrm{\mathbf{C}} \in \mathcal{R}^{1\times N}$ are the system matrices. Specifically, the matrix $\mathbf{A}$ represents the progression of the state vector\textit{ h}(\textit{t}). The projection matrix $\mathbf{B}$ specifies how much the input \(x\left(t\right) \) influences the hidden state. Through the projection parameter matrix $\mathbf{C}$, the hidden state is converted into the output. 

\par S4 model incorporates a timescale parameter, denoted as \(\Delta \), to discretize the continuous system parameters $\mathbf{A}$ and  $\mathbf{B}$ into their discrete equivalents $\overline{\mathrm{\mathbf{A}}}$ and $\overline{\mathrm{\mathbf{B}}}$. This process aims to adapt the continuous system into DL methods. This transformation typically employs the zero-order hold (ZOH) method that can be expressed as follows: 
\begin{equation}
\left\{
    \begin{aligned}
      &\overline{\mathbf{A}}=\exp (\Delta \mathbf{A})  \\
      &\overline{\mathbf{B}}=\left(\Delta \mathbf{A}\right)^{-1}\left(\exp \left(\Delta \mathbf{A}\right)-\mathbf{I}\right)\Delta \mathbf{B}
    \end{aligned}
\right.
\label{eq:ZOH}
\end{equation}

\par After the employment of ZOH method, the discrete-time SSM can be expressed as 
\begin{equation}
\left\{
    \begin{aligned}
      &h^{\prime}_{t} =\overline{\mathbf{A}} h_{t-1}+\overline{\mathbf{B}} x_{t}  \\
      &y_{t} =\mathbf{C} h_{t}
    \end{aligned}
\right.
\label{eq:discretation}
\end{equation}

\par Moreover, the Equation~\ref{eq:discretation} can be computed using a structured convolutional kernel $\overline{\mathbf{K}} \in \mathcal{R}^{\mathbf{L}}$. This process is expressed as
\begin{equation}
\left\{
    \begin{aligned}
      &\overline{\mathbf{K}}=\left(\mathbf{C} \overline{\mathbf{B}}, \mathbf{C} \overline{\mathbf{A}} \overline{\mathbf{B}}, \ldots, \mathbf{C}\overline{\mathbf{A}}^{\mathbf{L}-1} \overline{\mathbf{B}} \right)  \\
      &\mathbf{y}=\mathbf{x} * \overline{\mathbf{K}}
    \end{aligned}
\right.
\label{eq:global_conv}
\end{equation}
where $*$ denotes a convolutional operation. 

\begin{figure}
	\centering
	\includegraphics[scale=0.45]{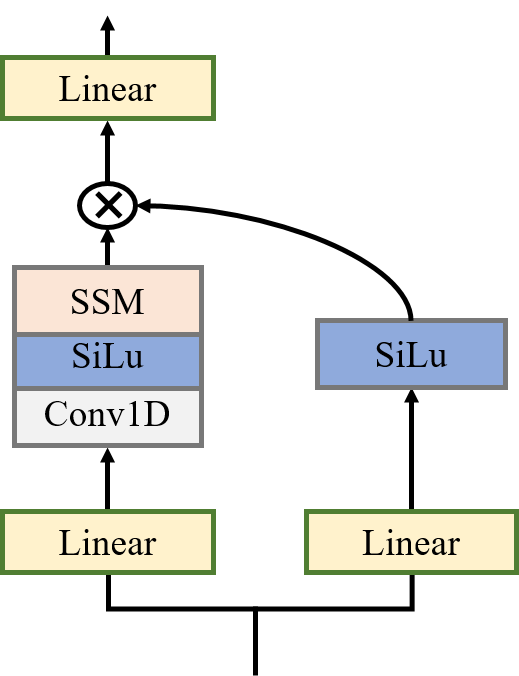}
	\caption{Structure of Mamba block.}
	\label{mamba}
\end{figure}

\subsection{Mamba Block}
As shown in Fig.~\ref{mamba}, the Mamba block integrates the mechanism of SSM with MLP, enabling the Mamba block to harness SSM's linear complexity and long-range dependencies learning ability, alongside the feature transformation capabilities inherent to MLPs. Being foundational to many SSM architectures, H3 block is integrated with to MLP block in the Mamba block. Mamba block includes two branches, which is similar to the Gated MLP (as shown in Fig.~\ref{architecture})\cite{gu2023mamba}. The main branch of Mamba block is derived from H3 block but the first multiplicative gate is replaced with a SiLU activation function. The second branch contains a Linear layer and a SiLu function. The outputs of these two branches are integrated with a multiplication computation. The resulting feature is then passed through a Linear layer to generate output.

\subsection{Temporal Prediction Branch}
After mamba block, the model generates the sequence $O_{n}=\{O^{0}_{n},...O^{L-1}_{n}\}$ for the $n^{th}$ image patch in a mini-batch. We hypothesize that this sequence, which contains embedded information, should also be able to predict the original time series $X_{n}=\{X^{0}_{n},...X^{L-1}_{n}\}$. Therefore, we define a task of predicting the original input sequence $X_{n}$ using $O_{n}$. Since the temporal length of the feature map is the same as the original data, we only need to adjust the number of channels in the feature map to match the channel dimension of the original data. Therefore, we use a Linear layer to transform $O_{n}\in \mathcal{R}^{L\times C_{1}}$ to $T_{n} \in \mathcal{R}^{L\times C}$. Besides, regarding that the earlier the position in the sequence, the more difficult it is to predict its value, we design a Positional Weight (PW) $\{\frac{1}{L},...,\frac{L}{L}\}$ that assigns larger weight with the increasing temporal order. This weight is applied to the MSE loss. The loss function of the temporal prediction branch is 
\begin{align} {\mathcal {L}}_{\text {tp}} (X,T) =  {\textstyle \sum_{1}^{N}} [\frac{1}{L}(X^{0}_{n}-T^{0}_{n})^2+\frac{2}{L}(X^{1}_{n}-T^{1}_{n})^2+...\notag\\
+\frac{L}{L}(X^{L-1}_{n}-T^{L-1}_{n})^2].\end{align}

\subsection{Overall Loss Function}
The overall loss function includes losses of these two branches. For the Classification Branch, we use 

 \begin{align}
\mathcal{L}_{cls}(Y,\hat{Y})=-\frac{1}{N}\sum_{n=1}^{N}\sum_{k=1}^{K}{Y_n^klog{\ \hat{Y}}_n^k}\
\end{align}

Where \textit{N }is the batch size, \textit{K }is the number of classes, $Y_n^k$ denotes the label of each class \textit{k} on the \textit{n\textsuperscript{th}} patch within the batch, and ${\ \hat{Y}}_n^k$ is the predicted probability of each class  on the \textit{n\textsuperscript{th}} patch.
\par To balance the two losses, we use a hyper-parameter $w_{0}$ to define the weight of $\mathcal {L}_{\text {tp}}$. We set $w_{0}$ as 0.03. We also use the ratio of the two losses, \(w_{1}=\frac{\mathcal{L}_{cls}}{\mathcal{L}_{tp}}\), to control the scale of $\mathcal {L}_{\text {tp}}$. The overall loss function is expressed as follows:

 \begin{align}
\mathcal{L}(Y,\hat{Y},X,T)=\mathcal{L}_{cls}+w_{0}w_{1}\mathcal{L}_{tp}.
\label{overallloss}\end{align}
\subsection{Architecture detail}
Serving as the spatial encoder, the ConvBlock consists of two Convolutional (Conv2D) layers, each applying a 3×3 convolutions with a padding of 1. This parameter setting preserves the spatial resolution of the input data. Both the Conv2D layers have 128 filters. Following each convolution is a batch normalization (BN) layer and a rectified linear unit (ReLU) activation function. The Mamba block is configured with its default parameter settings. For an input SITS data represented as $\mathrm{\mathbf{I}}\in \mathcal{R}^{T \times C \times H \times W}$, the ConvBlock applies convolutional calculation to each temporal frame, focusing on extracting the spatial features. The output feature of ConvBlock has the shape $T \times C_{1} \times H \times W$ where $C_{1}$ is 128 after the final convolution. This feature is then reshaped to $( H \times W) \times T \times C_{1} $, representing a set of $H \times W$ time series with length $T$ and $C_{1}$ channels. It is fed into the Mamba block to encode the temporal context and produce an output of the same shape. This output feature embedded with useful spatiotemporal information is then processed by two branches. For CBranch, a Max pooling layer is applied along the temporal dimension, generating a feature with the shape $H \times W\times C_{1} $. This feature is then fed into a ConvBlock, which is composed of a Conv2D layer that has $C_{cls}$ filters followed by a BN layer and a ReLU layer, where $C_{cls}$ is the number of crop categories. After this ConvBlock, the feature is decoded to generate a crop map with a shape of $H \times W\times C_{cls} $. For RBranch, we use a Linear layer to decode the feature from Mamba and predict the original SITS. The training and testing procedures are summarized in the Algorithm 1.

\begin{table}
\centering
 \setlength{\tabcolsep}{8mm}
 \renewcommand{\arraystretch}{2}
\begin{tabular}{l} 
\hline
\textbf{Algorithm 1 Training and Testing of SITSMamba}  \\ 
\hline

//\textit{Training}\\
\textbf{Input}: Image time series \(I\) and crop label \(Y\) in the training dataset.                                         \\
\textbf{Initialization}: Initialize parameters randomly.\\
for \(e\)= 0,1,2,...,\(E-1\) do\\
    \quad \(S\)=ConvBlock(\(I\)); \\
    \quad\(S_{rs}\)=Reshape(\(S\));\\
    \quad\(T\)=Mamba(\(S_{rs}\));  \\
    \quad
    \(T_{max}\)=Reshape(Maxpool(\(T\))); \\
     \quad\(O_{cls}\)=ConvBlock( \(T_{max}\));\\
     \quad \(O_{rec}\)=Linear(\(T\));\\
     \quad Calculate \(\mathcal{L}\) (\(Y\),\(O_{cls}\),\(I\),\(O_{rec}\)) in Eq.~\ref{overallloss}.\\
     \quad Update SITSMamba's parameters with Adam optimizer.\\
\textbf{end}\\
\textbf{Output}: Model weight of SITSMamba.\\
//\textit{Testing}\\
\textbf{Input}: Testing image time series \(I^{test}\)\\
Load the well-trained SITSMamba\\
\(S^{test}\)=Reshape(ConvBlock(\(I^{test}\)));\\
\(T^{test}\)=Mamba(\(S^{test}\));  \\
 \(T_{max}^{test}\)=Reshape(Maxpool(\(T^{test}\))); \\
\(O_{cls}^{test}\)=ConvBlock( \(T_{max}^{test}\));\\
\textbf{Output}: The prediction crop maps \(O_{cls}^{test}\).\\
\hline
\end{tabular}
\end{table}

\section{Experiments}\label{sec:4}
\subsection{Datasets}
We select two crop type datasets that covers more than 17 classes to evaluate the performance of the proposed method.

1) PASTIS Dataset \cite{garnot2021panoptic}:
Panoptic Agricultural Satellite TIme-Series (PASTIS) dataset is composed of four sub-regions in France. This dataset comprises 2,433 one-square-kilometer patches, spanning across four Sentinel-2 tile areas within France. Each patch is equipped with a series of multi-spectral satellite images, each featuring 10 bands and measured at a 10 m resolution of 128 × 128 pixels. This dataset contains 20 categories, which poses a challenge for classification task. The imagery was captured by Sentinel-2 satellites between September 2018 and November 2019, with the number of observations varying from 38 to 61 for each site. Leveraging the spatial and descriptive data of all parcels from the French Land Parcel Identification System (FLPIS), the PASTIS dataset annotates each image with corresponding semantic labels. The benchmark includes five folds, and we adopt fold-1 for training, fold-4 for validation and fold-5 for testing. PASTIS dataset features larger image size and longer temporal length. Therefore, we clip the image into a size of 32×32 pixels as the input for the models. We refer to this clipped dataset as PASTIS32. The number of pixels for each class in the PASTIS32 dataset is shown in Table~\ref{pastissta}. All the 20 classes are used for training and evaluation in this study. 

\begin{table}
\centering
 \renewcommand{\arraystretch}{1.3}
\caption{Statistics of classes in PASTIS32 dataset}
\label{pastissta}
\begin{tabular}{ccccc} 
\hline
Class number & Train   & Valid   & Test    & Class                        \\ 
\hline
0            & 3224570 & 3248735 & 3221643 & Background                   \\
1            & 1413557 & 1403832 & 1487831 & Meadow                       \\
2            & 532105  & 537363  & 580631  & Soft winter wheat            \\
3            & 731993  & 700052  & 620357  & Corn                         \\
4            & 155825  & 167928  & 171711  & Winter barley                \\
5            & 134868  & 152725  & 157965  & Winter rapeseed              \\
6            & 47192   & 56838   & 55308   & Spring barley                \\
7            & 74606   & 75896   & 90899   & Sunflower                    \\
8            & 241864  & 284057  & 219555  & Grapevine                    \\
9            & 56891   & 78380   & 67567   & Beet                         \\
10           & 51599   & 57181   & 53892   & Winter triticale             \\
11           & 94250   & 92010   & 88551   & Winter durum wheat           \\
12           & 89706   & 74056   & 81951   & \begin{tabular}[c]{@{}c@{}}Fruits, vegetables,\\~flowers\end{tabular}  \\
13           & 23529   & 28712   & 19712   & Potatoes                     \\
14           & 126927  & 130888  & 133936  & Leguminous fodder            \\
15           & 99748   & 87090   & 71771   & Soybeans                     \\
16           & 84323   & 73295   & 74429   & Orchard                      \\
17           & 43638   & 31477   & 49537   & Mixed cereal                 \\
18           & 27119   & 26993   & 53295   & Sorghum                      \\
19           & 724698  & 786188  & 825923  & Void label   \\
\hline
\end{tabular}
\end{table}

2) MTLCC Dataset \cite{russwurm2018multi}:
Our second dataset is the Multi-Temporal Land Cover Classification (MTLCC) benchmark. It encompasses an area of 102 km by 42 km in Munich, Germany. This dataset includes 18 classes. Comprising Sentinel-2 imagery from 2016, the dataset is organized into 6248 patches, each measuring 48 × 48 pixels with 13 spectral bands, and spanning a temporal dimension of 30 time steps. The training, validation, and test sets are split in the ratio of 4:1:1. The number of pixels for each crop type in the MTLCC dataset are shown in Table~\ref{mtlccsta}.

\begin{table}
\centering
 \renewcommand{\arraystretch}{1.3}
\caption{Statistics of classes in MTLCC dataset}
\label{mtlccsta}
\begin{tabular}{ccccc} 
\hline
Class number & Train   & Valid   & Test    & Class             \\ 
\hline
0            & 4332409 & 1332173 & 1298918 & Background        \\
1            & 63659   & 16366   & 21803   & Sugar beet        \\
2            & 46555   & 16125   & 20908   & Summer oat        \\
3            & 348652  & 103605  & 116779  & Meadow            \\
4            & 230050  & 64865   & 62283   & Rape              \\
5            & 144415  & 47226   & 26565   & Hop               \\
6            & 49901   & 12106   & 17087   & Winter spelt      \\
7            & 4332409 & 24251   & 27117   & Winter triticale  \\
8            & 63659   & 10233   & 11746   & Beans             \\
9            & 46555   & 7099    & 4452    & Peas              \\
10           & 348652  & 46072   & 57029   & Potatoes          \\
11           & 230050  & 8974    & 10434   & Soybeans          \\
12           & 144415  & 435     & 10559   & Asparagus         \\
13           & 49901   & 408749  & 383335  & Winter wheat      \\
14           & 474133  & 150344  & 119040  & Winter barley     \\
15           & 60616   & 12224   & 20969   & Winter rye        \\
16           & 165967  & 33170   & 33156   & Summer barley     \\
17           & 1388783 & 500735  & 437372  & Maize            \\
\hline
\end{tabular}
\end{table}
\subsection{Comparison with the SOTA methods}
We compare our approach against the cutting-edge and classical techniques for crop type mapping, which include a variety of methods that leverage CNN, RNN, and Transformer mechanisms for extracting temporal features. Among these methods, the spatial learning strategies under consideration include UNet, Fully Convolutional Network (FCN), and Atrous Convolution. For temporal encoding, we consider mechanisms based CNN, RNN, and Transformer. Our evaluation encompasses six SOTA methods: 3D U-Net \cite{m2019semantic}, BCLSTM\cite{russwurm2018multi}, UTAE\cite{garnot2021panoptic}, BUnetConvLSTM and BAtrousConvLSTM \cite{martinez2021fully}, and TSVIT\cite{tarasiou2023vits}.

The 3D U-Net, UTAE, and BUnetConvLSTM all utilize a U-shaped architecture designed for image semantic segmentation, featuring both downsampling and upsampling pathways. The 3D U-Net captures spatiotemporal data through three-dimensional convolutions, whereas UTAE and BUnetConvLSTM process temporal data using temporal attention mechanisms and Bi-directional ConvLSTM, respectively. BCLSTM incorporates FCN layers within its cells to account for spatial context, which can learn more spatial context compared with the standard LSTM models. The BAtrousConvLSTM integrates the features of the BCLSTM network with ASPP to capture multi-scale features effectively. While UTAE combines CNN and Transformer, TSVIT is a pure-Transformer architecture that adopts ViT architecture \cite{dosovitskiy2020image} to encode the spatial feature.

\subsection{Evaluation Metrics}
Three evaluation metrics, including overall accuracy (OA), intersection over union (IoU), and F1 score, are employed to evaluate the performance of the crop classification approaches. These evaluation metrics are computed as follows:

\[OA=\frac{TP+TN}{TP+FP+FN+TN}\]

\[IoU=\frac{TP}{TP+FP+FN}\]

\[F1\ Score=\frac{2TP}{2TP+FP+FN}\]

In these equations, TP, FP, and FN denote the numbers of pixels that are true-positive, false-positive, and false-negative for each crop category, respectively. To obtain a comprehensive assessment of the models' performances, we also report the average IoU (mIoU) and average F1 score (mF1). When calculating these two metrics, equal weight is assigned to every class to ensure that each crop type is considered equally.

\subsection{Implementation details}
On the PASTIS32 dataset, padding is adopted to ensure that the input SITS at different locations in a mini-batch have the same temporal length. As for MTLCC dataset, we sample 30 images from the SITS in the temporal dimension to ensure that each sample in a mini-batch has the same length in the temporal dimension. Given the variable input image sizes across datasets, we adjust the batch size to 8 for the PASTIS32 dataset, 4 for MTLCC datasets. All the models are trained for 100 epochs. The Adam optimizer is used to train the model. The learning rate is set to 0.0001. All runs are conducted on NVIDIA GeForce RTX 3090 GPU.

\subsection{Result}
\begin{figure*}
	\centering
	\includegraphics[scale=0.85]{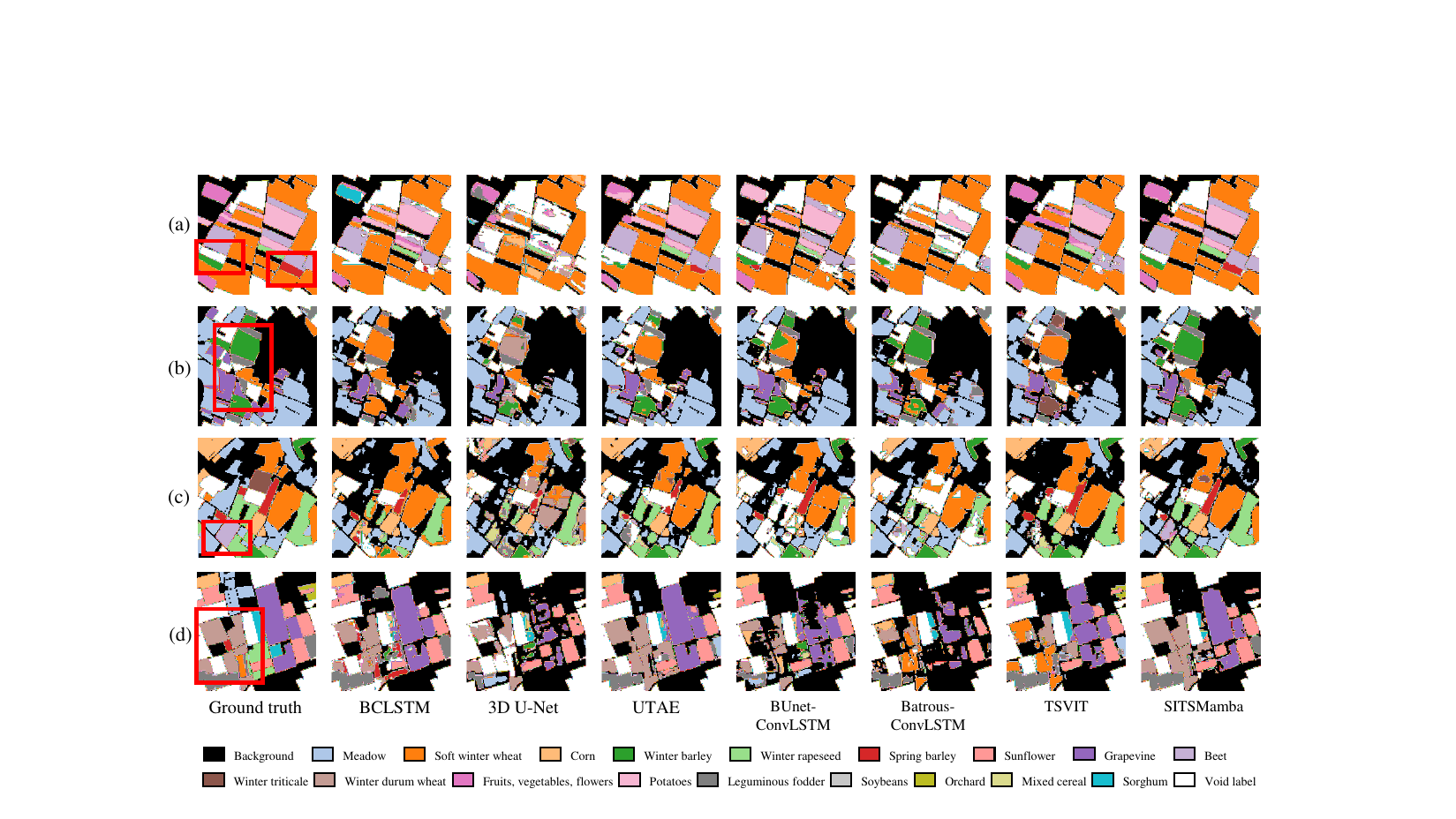}
	\caption{Qualitative evaluation of the models on PASTIS32 dataset.}
	\label{pastisview}
\end{figure*}
Table~\ref{pastisacc} and Table~\ref{mtlccacc} show the quantitative comparison results on PASTIS32 dataset and MTLCC dataset, respectively. As shown in Table~\ref{pastisacc}, SITSMamba outperforms other methods in all the three metrics. Among the other methods, UTAE achieves the best result. TSVIT achieves a result slightly lower than UTAE but outperforms the other methods that use LSTM as the temporal encoder. Apart from the overall evaluation, we also show the F1 score on all the crop types in Table ~\ref{pastisf1}. SITSMamba obtains the highest accuracy on most classes, especially for those classes hard to classify, such as class 6, 10, 12, and 18. 

\begin{table}
\centering
 \setlength{\tabcolsep}{5mm}
 \renewcommand{\arraystretch}{1.2}
\caption{Quantitative comparison results on PASTIS32 Dataset.}
\label{pastisacc}
\begin{tabular}{lccc} 
\hline
Method          & OA     & mIoU   & mF1     \\ 
\hline
BCLSTM          & 0.6697 & 0.3057 & 0.4282  \\
3D U-Net        & 0.7057 & 0.3999 & 0.5358  \\
UTAE            & 0.7349 & 0.4686 & 0.6107  \\
BUnetConvLSTM   & 0.7155 & 0.4025 & 0.5401  \\
BAtrousConvLSTM & 0.7024 & 0.3917 & 0.5327  \\
TSVIT           & 0.7200 & 0.4561 & 0.5999  \\
SITSMamba       & \textbf{0.7416} & \textbf{0.5005} & \textbf{0.6449}  \\
\hline

\end{tabular}
\end{table}

\begin{table}
\centering
 \setlength{\tabcolsep}{5mm}
 \renewcommand{\arraystretch}{1.2}
\caption{Quantitative comparison results on MTLCC Dataset.}
\begin{tabular}{lccc} 
\hline
Method& OA& mIoU & mF1\\ 
\hline
BCLSTM& 0.8829& 0.6353& 0.7449 \\
3D U-Net        & 0.8926& 0.6891& 0.7908 \\
UTAE  & 0.9063& 0.7467& 0.8391 \\
BUnetConvLSTM   & 0.9079& 0.7399& 0.8344 \\
BAtrousConvLSTM & 0.9010& 0.7328& 0.8262 \\
TSVIT & 0.8954& 0.7358& 0.8299 \\
SITSMamba       & \textbf{0.9104} & \textbf{0.7639} & \textbf{0.8496}  \\
\hline
\label{mtlccacc}
\end{tabular}
\end{table}

As shown in Table~\ref{mtlccacc}, SITSMamba shows the best performance among the comparative methods on MTLCC dataset. Following SITSMamba, UTAE achieves the second-best result in mIoU and mF1. Among the LSTM-based methods, BUnetConvLSTM obtains the best result, which has the slightly lower accuracy than UTAE in mIoU and mF1 metrics. Additionally, the F1 score of all crop types are shown on MTLCC dataset (in Table~\ref{mtlccf1}). SITSMamba achieves the highest F1 score in 9 out of the 17 crop classes,  while other methods outperform it in fewer classes.

\begin{table*}
\centering \renewcommand{\arraystretch}{1.2}
 \setlength{\tabcolsep}{4.5mm}
\caption{F1 score of each crop type on PASTIS32 Dataset.}
\label{pastisf1}
\begin{tabular}{cccccccc}
\hline
Class & BCLSTM & 3D U-Net & UTAE            & BUnetConvLSTM & BAtrousConvLSTM & TSVIT  & SITSMamba        \\
\hline
1     & 0.6849 & 0.7008   & \textbf{0.7235} & 0.7221        & 0.7015          & 0.7037 & 0.7202           \\
2     & 0.7525 & 0.7756   & 0.8406          & 0.8149        & 0.8108          & 0.8277 & \textbf{0.8556}  \\
3     & 0.7702 & 0.8222   & 0.8521          & 0.8240        & 0.8099          & 0.8349 & \textbf{0.8566}  \\
4     & 0.5095 & 0.5800   & 0.7720          & 0.7360        & 0.7549          & 0.7595 & \textbf{0.7783}  \\
5     & 0.8066 & 0.8486   & 0.8521          & 0.8190        & 0.8272          & 0.8623 & \textbf{0.8648}  \\
6     & 0.2579 & 0.4263   & 0.5854          & 0.4262        & 0.4266          & 0.5739 & \textbf{0.5962}  \\
7     & 0.5440 & 0.6488   & 0.6484          & 0.6296        & 0.5848          & 0.6613 & \textbf{0.7493}  \\
8     & 0.5813 & 0.6436   & 0.6559          & 0.6075        & 0.5185          & 0.6323 & \textbf{0.6791}  \\
9     & 0.1918 & 0.7098   & \textbf{0.8314} & 0.7642        & 0.5635          & 0.7583 & 0.8198           \\
10    & 0.1629 & 0.0461   & 0.3407          & 0.3310        & 0.2777          & 0.1961 & \textbf{0.4632}           \\
11    & 0.4362 & 0.4906   & \textbf{0.6480} & 0.4639        & 0.5252          & 0.6440 & 0.6297           \\
12    & 0.2508 & 0.3453   & 0.3712          & 0.3720        & 0.2411          & 0.4228 & \textbf{0.4730}  \\
13    & 0.0685 & 0.5966   & 0.5045          & 0.4967        & 0.3669          & 0.4693 & \textbf{0.6007}  \\
14    & 0.2873 & 0.3333   & 0.4773          & 0.3700        & 0.3930          & 0.4622 & \textbf{0.5248}  \\
15    & 0.6447 & 0.8066   & 0.8190          & 0.7171        & 0.7901          & 0.8299 & \textbf{0.8516}  \\
16    & 0.1421 & 0.4223   & \textbf{0.5164} & 0.1885        & 0.3828          & 0.5016 & 0.5055           \\
17    & 0.3177 & 0.2844   & \textbf{0.4527} & 0.3533        & 0.3370          & 0.4089 & 0.4502           \\
18    & 0.1545 & 0.1742   & 0.2697          & 0.1219        & 0.2924          & 0.3891 & \textbf{0.4139}\\ \hline
\end{tabular}
\end{table*}

\begin{table*}
\centering \renewcommand{\arraystretch}{1.2}
 \setlength{\tabcolsep}{4.5mm}
\caption{F1 score of each crop type on MTLCC Dataset.}
\label{mtlccf1}
\begin{tabular}{cccccccc}
\hline
Class & BCLSTM & 3D U-Net & UTAE            & BUnetConvLSTM   & BAtrousConvLSTM & TSVIT  & SITSMamba        \\
\hline
1     & 0.8733 & 0.8874   & 0.8908          & 0.9042          & \textbf{0.9356} & 0.9028 & 0.9193           \\
2     & 0.6727 & 0.7247   & 0.8041          & \textbf{0.8236} & 0.8077          & 0.8032 & 0.8151           \\
3     & 0.1586 & 0.3136   & 0.3331          & \textbf{0.3458} & 0.3342          & 0.3167 & 0.3082           \\
4     & 0.9424 & 0.9518   & 0.9562          & 0.9539          & \textbf{0.9639} & 0.9529 & 0.9629           \\
5     & 0.8983 & 0.9222   & 0.9369          & 0.9334          & \textbf{0.9381} & 0.9357 & 0.9371           \\
6     & 0.4854 & 0.5897   & \textbf{0.7332} & 0.7242          & 0.6573          & 0.6564 & 0.7313           \\
7     & 0.3861 & 0.4192   & 0.5978          & 0.5812          & 0.4853          & 0.5349 & \textbf{0.6263}  \\
8     & 0.8928 & 0.9119   & 0.9202          & 0.9195          & 0.9261          & 0.9008 & \textbf{0.9378}  \\
9     & 0.6941 & 0.7362   & 0.7885          & 0.7338          & 0.8004          & 0.8453 & \textbf{0.8588}  \\
10    & 0.8518 & 0.8823   & 0.8840          & 0.8541          & 0.8883          & 0.8874 & \textbf{0.9066}  \\
11    & 0.7452 & 0.8860   & 0.8948          & 0.8709          & 0.8732          & 0.8919 & \textbf{0.9095}  \\
12    & 0.8153 & 0.8831   & 0.9071          & 0.8967          & 0.8686          & 0.9032 & \textbf{0.9190}  \\
13    & 0.9146 & 0.9226   & 0.9347          & 0.9377          & 0.9344          & 0.9248 & \textbf{0.9389}  \\
14    & 0.9163 & 0.9334   & 0.9416          & 0.9442          & 0.9420          & 0.9310 & \textbf{0.9480}  \\
15    & 0.4817 & 0.5468   & 0.8005          & \textbf{0.8133} & 0.7482          & 0.7973 & 0.7941           \\
16    & 0.8228 & 0.8548   & \textbf{0.8983} & 0.8981          & 0.8918          & 0.8925 & 0.8927           \\
17    & 0.9455 & 0.9532   & 0.9600          & 0.9587          & 0.9570          & 0.9482 & \textbf{0.9613} \\
\hline
\end{tabular}
\end{table*}

\begin{figure*}
	\centering
	\includegraphics[scale=0.85]{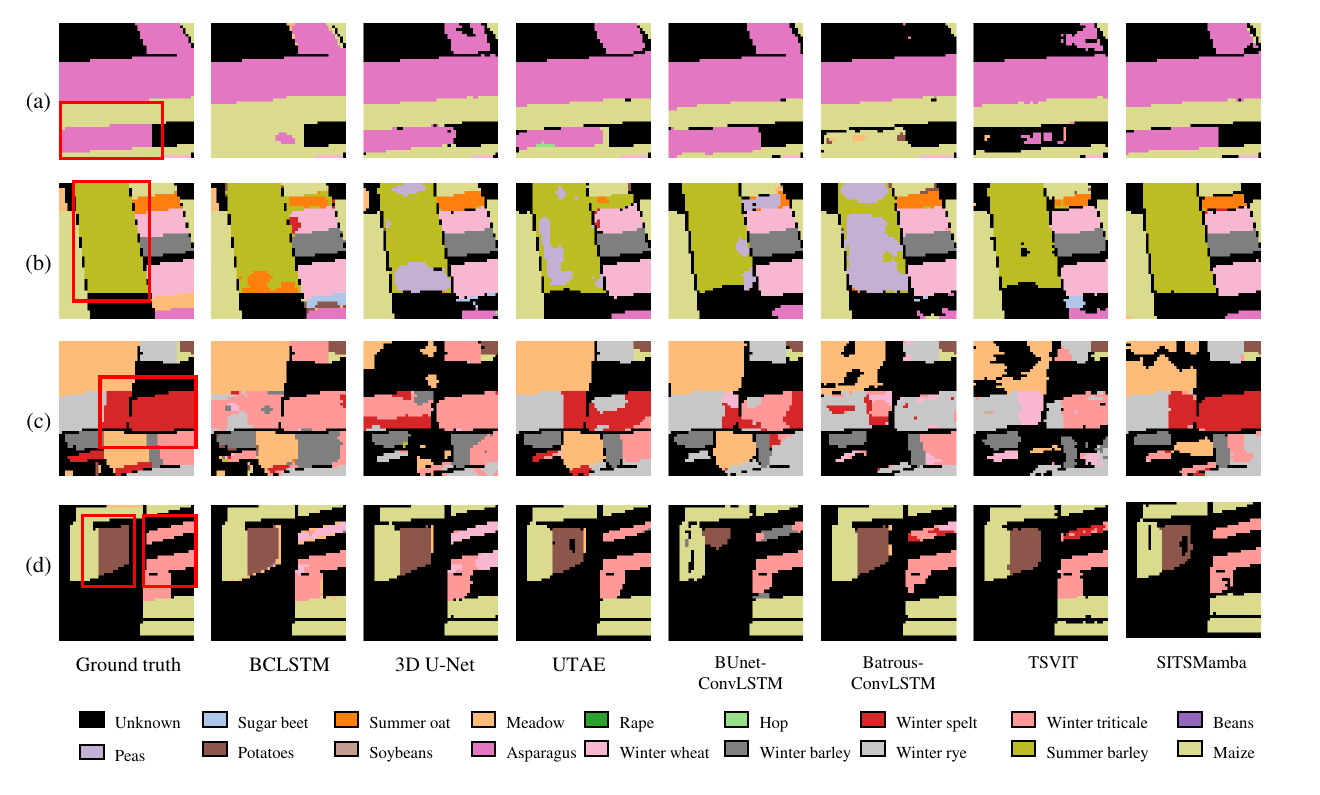}
	\caption{Qualitative evaluation of the models on MTLCC dataset.}
	\label{mtlccview}
\end{figure*}

We present the crop mapping results of the competing methods on the PASTIS32 dataset in Fig.~\ref{pastisview}. In areas with complex field shapes and fragmented parcels, SITSMamba has better prediction results than the competing methods. As shown in the red rectangles, SITS can accurately predict various parcels, including the winter barley, spring barley, beet, and winter durum wheat. Among the competing methods, BCLSTM can accurately predict the crop types of a parcel, but it often fails to delineate the correct boundary. 3D U-Net generates a more fragile result, which is obvious in the soft winter wheat in row (c) and the grapevine parcel in row (d). This may be due to its insufficient temporal encoding using CNN. The other LSTM-based methods, BUnetConvLSTM and BAtrousConvLSTM, also generate many tiny wrong parcels. On the contrary, the Transformer-based methods, that is, UTAE and TSVIT, produce better prediction maps. This result is consistent with the quantitative comparison in Table~\ref{pastisacc} that the Transformer-based methods surpass the LSTM-based methods. 

Fig.~\ref{mtlccview} presents a visual comparison of various models on the MTLCC dataset. SITSMamba outperforms the other methods by correctly maintaining the boundary of the asparagus parcel, as shown in row (a). Unlike the comparative models that produce the parcel with incomplete extent or with holes, SITSMamba predicts the whole summer barley parcel as shown in row (b). Additionally, SITSMamba generates the winter spelt more accurately in row (c), whereas the other methods tend to misclassify it to winter triticale or winter rye. However, there are instances where SITSMamba does not outperform other methods in classification accuracy, such as with the potatoes and winter triticale areas in row (d). Nonetheless, SITSMamba generally demonstrates the ability to produce accurate crop maps.

\subsection{Ablation study}
Table~\ref{ablationpastis} shows the ablation study of the effectiveness of RBranch, PW and $w_{1}$. Excluding the use of PW, which adjusts the weight of RBranch based on temporal position, our model's performance reduces by 0.74\% in mF1. This result suggests that PW's emphasis on later temporal positions can indeed enhance model performance. Additionally, the model's accuracy declines without the application of $w_{1}$, which is designed to balance the training between CBranch and RBranch. Furthermore, omitting RBranch as an auxiliary branch during training results in poorer performance compared to the model trained with both branches, thus highlighting the beneficial role of RBranch.  

\begin{table}
\centering
  \renewcommand{\arraystretch}{1.2}
  \setlength{\tabcolsep}{5mm}
\caption{Ablation analysis of the RBranch and \protect\\ the proposed weighting strategy.}
\label{ablationpastis}
\begin{tabular}{lccc} 
\hline
Method    & OA     & mIoU   & mF1     \\ 
\hline
SITSMamba & \textbf{0.7416} & \textbf{0.5005} & \textbf{0.6449}  \\
W/o PW    & 0.7406 & 0.4970 & 0.6401  \\
W/o w1    & 0.7351 & 0.4828 & 0.6250  \\
W/o RBranch    & 0.7337 & 0.4875 & 0.6301  \\
\hline
\end{tabular}
\end{table}

\begin{figure*}
	\centering
	\includegraphics[scale=1.]{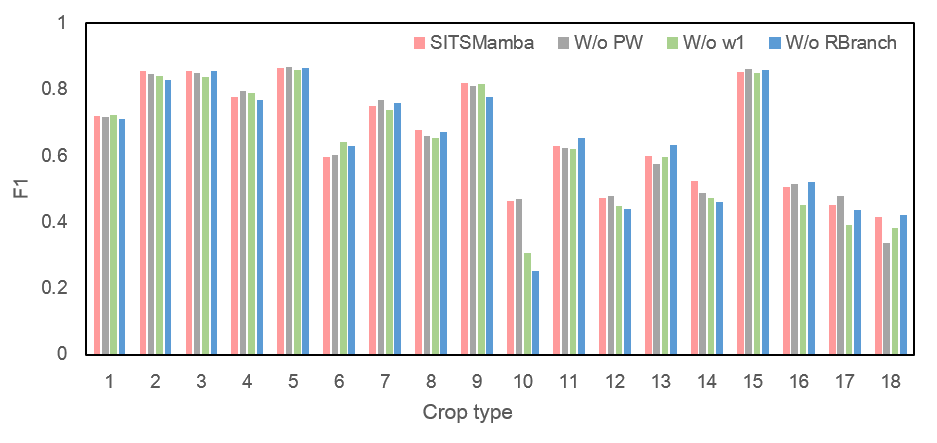}
	\caption{F1 scores of crop types with different methods on PASTIS32 dataset.}
	\label{ablationF1pastis}
\end{figure*}

Fig.~\ref{ablationF1pastis} details the F1 scores for various crop types across different methods in the ablation study. The F1 scores across the crop types show a similar trend for the four methods. Among them, the method that does not incorporate PW closely matches SITSMamba's performance and surpasses it in a few classes. The method without using $w_{1}$ obtains inferior F1 scores in most classes compared with SITSMamba, particularly in the more challenging ones, such as types 10, 12, 14, 17, and 18. This decline in most classes highlights the importance of $w_{1}$, which is designed to balance the training between CBranch and RBranch. The method without using RBranch shows lower accuracy in most classes when compared with SITSMamba and the version without PW. Especially, for the most challenging class (type 10), the method without using RBranch obtains the lowest F1 score. This result demonstrates the effectiveness of RBranch, along with the integration of PW and $w_{1}$, in tackling the complex task of crop classification.
\par
\begin{figure}
	\centering
	\includegraphics[scale=0.75]{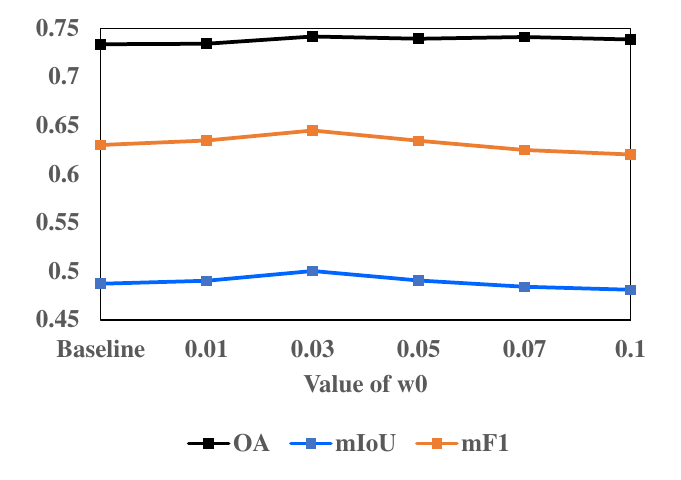}
	\caption{Parameter analysis for the choice of $w_{0}$.}
	\label{Parameter}
\end{figure}

 Furthermore, the parameter $w_{0}$ controls the extent of RBranch's influence on the training process. It is crucial as setting it too low results in insufficient learning of auxiliary information from RBranch, while setting it too high can disrupt the CBranch's training. Fig.~\ref{Parameter} presents the performance of SITSMamba across a range of $w_{0}$ values, using a baseline that lacks RBranch. As shown in  Fig.~\ref{Parameter}, OA, mIoU and mF1 all improve with the increasing $w_{0}$. When $w_{0}$ is equal to 0.03, the model attains the highest scores in all the three metrics, suggesting that this value effectively balances the contributions of CBranch and RBranch. However, further increases in $w_{0}$ lead to a decline in SITSMamba's performance, suggesting that excessive focus on RBranch's training can impede the crop type classification task. Therefore, we choose $w_{0}$=0.03 as the default setting in SITSMamba.

\section{Discussion}\label{sec:5}

\subsection{Complexity analysis}
Table~\ref{params1} lists the number of parameters and GFLOPs of comparative models. SITS has the least number of parameter, which is similar to that of BCLSTM. 3D U-Net has the largest number of parameters because it uses multiple 3-D convolutional layers to learn spatiotemporal representations. Among the Transformer-based methods, UTAE has a smaller number of parameters and also lower GFLOPs. This is because it simplifies the self-attention mechanism to reduce the number of parameter and the corresponding computational requirements. SITSMamba has larger GFLOPs compared to UTAE and BUnetConvLSTM, probably because these two comparative methods employ UNet to downsample the image size, which reduces the computational load in temporal encoding. Therefore, we would like to explore the more efficient spatiotemporal encoding methods to further reduce the complexity of SITSMamba.

\begin{table}\renewcommand{\arraystretch}{1.2}
\centering
\setlength{\tabcolsep}{5mm}
\caption{Comparison of SITS classification models \protect\\ in computational cost}
\begin{tabular}{lcc}

\hline
Method& \multicolumn{1}{l}{Params (M)} & \multicolumn{1}{l}{GFLOPs}    \\ \hline
BCLSTM          & {0.29}    & {91.80}  \\
3D U-Net        & {6.18}    & {273.99} \\
UTAE            & {1.09}    & {\textbf{28.77}}  \\
BUnetConvLSTM   & {1.41}    & {29.48}  \\
BAtrousConvLSTM & {2.24}    & {356.88} \\
TSVIT           & {2.16}    & {84.35}  \\
SITSMamba       & {\textbf{0.25}}    & {56.53}  \\ \hline
\end{tabular}\label{params1}
\end{table}

\subsection{Limitation and Future Work}
Although our method surpasses the previous SOTA approaches, it still has some
limitations and can be further improved. Firstly, the observations within SITS can be affected by anomalous weather conditions, such as rain, snow, or cloud cover, which may introduce abnormal values in certain locations or at specific times. In the RBranch, we have designed a task to reconstruct the original SITS, including these potentially noisy values. These anomalies could potentially impair the model's optimization process through the training of RBranch. Therefore, incorporating cloud masks or snow masks could further enhance SITSMamba's performance by filtering out such noise.

Secondly, while temporal information is pivotal for crop classification, our current implementation of Mamba focuses solely on the temporal dimension. Given Mamba's demonstrated capability for analyzing very high resolution (VHR) and hyperspectral images, its potential to learn spatial information from 10-meter resolution Sentinel-2 SITS data remains unexplored. Therefore, we intend to investigate the incorporation of Mamba's spatial learning capabilities to fully exploit the spatial-temporal characteristics of SITS data. This could lead to more robust and accurate crop classification models that better utilize the rich information content available in multi-spectral satellite imagery.

\section{Conclusion}\label{sec:6}
In this study, we present a method called SITSMamba, which is specifically designed for crop classification using remote sensing image time series. Distinct from previous approaches that merely used crop label information for supervised training, our SITSMamba contains multi-task training with two branches of decoder. One branch (CBranch) aims for crop classification, and the other branch (RBranch) constructs the original SITS. With the auxiliary training of RBranch, SITSMamba learns more valuable latent knowledge from SITS data that enhances the crop classification. Furthermore, we design a PW to assign an increasing weight across time to the time series when calculating the loss of RBranch. Experiments conducted on datasets containing nearly 20 crop types demonstrate SITSMamba's superior performance in crop classification compared to existing SOTA techniques. The ablation study confirms the critical role of RBranch in enhancing the model's performance, highlighting the benefits of reconstructing the original SITS data alongside the classification task.

\bibliographystyle{IEEEtran}
\bibliography{ref}

\begin{thebibliography}{10}
\providecommand{\url}[1]{#1}
\csname url@samestyle\endcsname
\providecommand{\newblock}{\relax}
\providecommand{\bibinfo}[2]{#2}
\providecommand{\BIBentrySTDinterwordspacing}{\spaceskip=0pt\relax}
\providecommand{\BIBentryALTinterwordstretchfactor}{4}
\providecommand{\BIBentryALTinterwordspacing}{\spaceskip=\fontdimen2\font plus
\BIBentryALTinterwordstretchfactor\fontdimen3\font minus \fontdimen4\font\relax}
\providecommand{\BIBforeignlanguage}[2]{{%
\expandafter\ifx\csname l@#1\endcsname\relax
\typeout{** WARNING: IEEEtran.bst: No hyphenation pattern has been}%
\typeout{** loaded for the language `#1'. Using the pattern for}%
\typeout{** the default language instead.}%
\else
\language=\csname l@#1\endcsname
\fi
#2}}
\providecommand{\BIBdecl}{\relax}
\BIBdecl

\bibitem{lin2022early}
C.~Lin, L.~Zhong, X.-P. Song, J.~Dong, D.~B. Lobell, and Z.~Jin, ``Early-and in-season crop type mapping without current-year ground truth: Generating labels from historical information via a topology-based approach,'' \emph{Remote Sensing of Environment}, vol. 274, p. 112994, 2022.

\bibitem{wolanin2019estimating}
A.~Wolanin, G.~Camps-Valls, L.~G{\'o}mez-Chova, G.~Mateo-Garc{\'\i}a, C.~van~der Tol, Y.~Zhang, and L.~Guanter, ``Estimating crop primary productivity with sentinel-2 and landsat 8 using machine learning methods trained with radiative transfer simulations,'' \emph{Remote sensing of environment}, vol. 225, pp. 441--457, 2019.

\bibitem{yuan2023bridging}
Y.~Yuan, L.~Lin, Z.-G. Zhou, H.~Jiang, and Q.~Liu, ``Bridging optical and sar satellite image time series via contrastive feature extraction for crop classification,'' \emph{ISPRS Journal of Photogrammetry and Remote Sensing}, vol. 195, pp. 222--232, 2023.

\bibitem{cole2018science}
M.~B. Cole, M.~A. Augustin, M.~J. Robertson, and J.~M. Manners, ``The science of food security,'' \emph{npj Science of Food}, vol.~2, no.~1, p.~14, 2018.

\bibitem{xu2021towards}
J.~Xu, J.~Yang, X.~Xiong, H.~Li, J.~Huang, K.~Ting, Y.~Ying, and T.~Lin, ``Towards interpreting multi-temporal deep learning models in crop mapping,'' \emph{Remote Sensing of Environment}, vol. 264, p. 112599, 2021.

\bibitem{fan2014characterizing}
C.~Fan, B.~Zheng, S.~W. Myint, and R.~Aggarwal, ``Characterizing changes in cropping patterns using sequential landsat imagery: An adaptive threshold approach and application to phoenix, arizona,'' \emph{International Journal of Remote Sensing}, vol.~35, no.~20, pp. 7263--7278, 2014.

\bibitem{foerster2012crop}
S.~Foerster, K.~Kaden, M.~Foerster, and S.~Itzerott, ``Crop type mapping using spectral--temporal profiles and phenological information,'' \emph{Computers and Electronics in Agriculture}, vol.~89, pp. 30--40, 2012.

\bibitem{sakamoto2010two}
T.~Sakamoto, B.~D. Wardlow, A.~A. Gitelson, S.~B. Verma, A.~E. Suyker, and T.~J. Arkebauer, ``A two-step filtering approach for detecting maize and soybean phenology with time-series modis data,'' \emph{Remote Sensing of Environment}, vol. 114, no.~10, pp. 2146--2159, 2010.

\bibitem{pelletier2016assessing}
C.~Pelletier, S.~Valero, J.~Inglada, N.~Champion, and G.~Dedieu, ``Assessing the robustness of random forests to map land cover with high resolution satellite image time series over large areas,'' \emph{Remote Sensing of Environment}, vol. 187, pp. 156--168, 2016.

\bibitem{vaswani2017attention}
A.~Vaswani, N.~Shazeer, N.~Parmar, J.~Uszkoreit, L.~Jones, A.~N. Gomez, {\L}.~Kaiser, and I.~Polosukhin, ``Attention is all you need,'' \emph{Advances in neural information processing systems}, vol.~30, 2017.

\bibitem{ienco2017land}
D.~Ienco, R.~Gaetano, C.~Dupaquier, and P.~Maurel, ``Land cover classification via multitemporal spatial data by deep recurrent neural networks,'' \emph{IEEE Geoscience and Remote Sensing Letters}, vol.~14, no.~10, pp. 1685--1689, 2017.

\bibitem{yuan2020using}
Y.~Yuan, L.~Lin, L.-Z. Huo, Y.-L. Kong, Z.-G. Zhou, B.~Wu, and Y.~Jia, ``Using an attention-based lstm encoder--decoder network for near real-time disturbance detection,'' \emph{IEEE Journal of Selected Topics in Applied Earth Observations and Remote Sensing}, vol.~13, pp. 1819--1832, 2020.

\bibitem{russwurm2017temporal}
M.~Ru{\ss}wurm and M.~Korner, ``Temporal vegetation modelling using long short-term memory networks for crop identification from medium-resolution multi-spectral satellite images,'' in \emph{Proceedings of the IEEE conference on computer vision and pattern recognition workshops}, 2017, pp. 11--19.

\bibitem{zhong2019deep}
L.~Zhong, L.~Hu, and H.~Zhou, ``Deep learning based multi-temporal crop classification,'' \emph{Remote sensing of environment}, vol. 221, pp. 430--443, 2019.

\bibitem{russwurm2020self}
M.~Ru{\ss}wurm and M.~K{\"o}rner, ``Self-attention for raw optical satellite time series classification,'' \emph{ISPRS journal of photogrammetry and remote sensing}, vol. 169, pp. 421--435, 2020.

\bibitem{garnot2021panoptic}
V.~S.~F. Garnot and L.~Landrieu, ``Panoptic segmentation of satellite image time series with convolutional temporal attention networks,'' in \emph{Proceedings of the IEEE/CVF International Conference on Computer Vision}, 2021, pp. 4872--4881.

\bibitem{yuan2020self}
Y.~Yuan and L.~Lin, ``Self-supervised pretraining of transformers for satellite image time series classification,'' \emph{IEEE Journal of Selected Topics in Applied Earth Observations and Remote Sensing}, vol.~14, pp. 474--487, 2020.

\bibitem{smith2022simplified}
J.~T. Smith, A.~Warrington, and S.~W. Linderman, ``Simplified state space layers for sequence modeling,'' \emph{arXiv preprint arXiv:2208.04933}, 2022.

\bibitem{ma2022mega}
X.~Ma, C.~Zhou, X.~Kong, J.~He, L.~Gui, G.~Neubig, J.~May, and L.~Zettlemoyer, ``Mega: moving average equipped gated attention,'' \emph{arXiv preprint arXiv:2209.10655}, 2022.

\bibitem{gu2023mamba}
A.~Gu and T.~Dao, ``Mamba: Linear-time sequence modeling with selective state spaces,'' \emph{arXiv preprint arXiv:2312.00752}, 2023.

\bibitem{russwurm2018multi}
M.~Ru{\ss}wurm and M.~K{\"o}rner, ``Multi-temporal land cover classification with sequential recurrent encoders,'' \emph{ISPRS International Journal of Geo-Information}, vol.~7, no.~4, p. 129, 2018.

\bibitem{yuan2022sits}
Y.~Yuan, L.~Lin, Q.~Liu, R.~Hang, and Z.-G. Zhou, ``Sits-former: A pre-trained spatio-spectral-temporal representation model for sentinel-2 time series classification,'' \emph{International Journal of Applied Earth Observation and Geoinformation}, vol. 106, p. 102651, 2022.

\bibitem{dumeur2024self}
I.~Dumeur, S.~Valero, and J.~Inglada, ``Self-supervised spatio-temporal representation learning of satellite image time series,'' \emph{IEEE Journal of Selected Topics in Applied Earth Observations and Remote Sensing}, 2024.

\bibitem{skakun2017early}
S.~Skakun, B.~Franch, E.~Vermote, J.-C. Roger, I.~Becker-Reshef, C.~Justice, and N.~Kussul, ``Early season large-area winter crop mapping using modis ndvi data, growing degree days information and a gaussian mixture model,'' \emph{Remote Sensing of Environment}, vol. 195, pp. 244--258, 2017.

\bibitem{zhao2020robust}
J.~Zhao, Y.~Zhong, X.~Hu, L.~Wei, and L.~Zhang, ``A robust spectral-spatial approach to identifying heterogeneous crops using remote sensing imagery with high spectral and spatial resolutions,'' \emph{Remote Sensing of Environment}, vol. 239, p. 111605, 2020.

\bibitem{li2023development}
H.~Li, X.-P. Song, M.~C. Hansen, I.~Becker-Reshef, B.~Adusei, J.~Pickering, L.~Wang, L.~Wang, Z.~Lin, V.~Zalles \emph{et~al.}, ``Development of a 10-m resolution maize and soybean map over china: Matching satellite-based crop classification with sample-based area estimation,'' \emph{Remote Sensing of Environment}, vol. 294, p. 113623, 2023.

\bibitem{pelletier2019temporal}
C.~Pelletier, G.~I. Webb, and F.~Petitjean, ``Temporal convolutional neural network for the classification of satellite image time series,'' \emph{Remote Sensing}, vol.~11, no.~5, p. 523, 2019.

\bibitem{garnot2019time}
V.~S.~F. Garnot, L.~Landrieu, S.~Giordano, and N.~Chehata, ``Time-space tradeoff in deep learning models for crop classification on satellite multi-spectral image time series,'' in \emph{IGARSS 2019-2019 IEEE International Geoscience and Remote Sensing Symposium}.\hskip 1em plus 0.5em minus 0.4em\relax IEEE, 2019, pp. 6247--6250.

\bibitem{garnot2020satellite}
{Garnot, Vivien Sainte Fare and Landrieu, Loic and Giordano, Sebastien and Chehata, Nesrine}, ``Satellite image time series classification with pixel-set encoders and temporal self-attention,'' in \emph{Proceedings of the IEEE/CVF Conference on Computer Vision and Pattern Recognition}, 2020, pp. 12\,325--12\,334.

\bibitem{tarasiou2023vits}
M.~Tarasiou, E.~Chavez, and S.~Zafeiriou, ``Vits for sits: Vision transformers for satellite image time series,'' in \emph{Proceedings of the IEEE/CVF Conference on Computer Vision and Pattern Recognition}, 2023, pp. 10\,418--10\,428.

\bibitem{mehta2022long}
H.~Mehta, A.~Gupta, A.~Cutkosky, and B.~Neyshabur, ``Long range language modeling via gated state spaces,'' \emph{arXiv preprint arXiv:2206.13947}, 2022.

\bibitem{wang2023selective}
J.~Wang, W.~Zhu, P.~Wang, X.~Yu, L.~Liu, M.~Omar, and R.~Hamid, ``Selective structured state-spaces for long-form video understanding,'' in \emph{Proceedings of the IEEE/CVF Conference on Computer Vision and Pattern Recognition}, 2023, pp. 6387--6397.

\bibitem{chen2024changemamba}
H.~Chen, J.~Song, C.~Han, J.~Xia, and N.~Yokoya, ``Changemamba: Remote sensing change detection with spatio-temporal state space model,'' \emph{arXiv preprint arXiv:2404.03425}, 2024.

\bibitem{zhao2024rs}
S.~Zhao, H.~Chen, X.~Zhang, P.~Xiao, L.~Bai, and W.~Ouyang, ``Rs-mamba for large remote sensing image dense prediction,'' \emph{arXiv preprint arXiv:2404.02668}, 2024.

\bibitem{he20243dss}
Y.~He, B.~Tu, B.~Liu, J.~Li, and A.~Plaza, ``3dss-mamba: 3d-spectral-spatial mamba for hyperspectral image classification,'' \emph{arXiv preprint arXiv:2405.12487}, 2024.

\bibitem{yao2024spectralmamba}
J.~Yao, D.~Hong, C.~Li, and J.~Chanussot, ``Spectralmamba: Efficient mamba for hyperspectral image classification,'' \emph{arXiv preprint arXiv:2404.08489}, 2024.

\bibitem{li2024mambahsi}
Y.~Li, Y.~Luo, L.~Zhang, Z.~Wang, and B.~Du, ``Mambahsi: Spatial-spectral mamba for hyperspectral image classification,'' \emph{IEEE Transactions on Geoscience and Remote Sensing}, 2024.

\bibitem{fu2024ssumamba}
G.~Fu, F.~Xiong, J.~Lu, and J.~Zhou, ``Ssumamba: Spatial-spectral selective state space model for hyperspectral image denoising,'' \emph{IEEE Transactions on Geoscience and Remote Sensing}, 2024.

\bibitem{m2019semantic}
R.~M~Rustowicz, R.~Cheong, L.~Wang, S.~Ermon, M.~Burke, and D.~Lobell, ``Semantic segmentation of crop type in africa: A novel dataset and analysis of deep learning methods,'' in \emph{Proceedings of the IEEE/cvf conference on computer vision and pattern recognition workshops}, 2019, pp. 75--82.

\bibitem{martinez2021fully}
J.~A.~C. Martinez, L.~E.~C. La~Rosa, R.~Q. Feitosa, I.~D. Sanches, and P.~N. Happ, ``Fully convolutional recurrent networks for multidate crop recognition from multitemporal image sequences,'' \emph{ISPRS Journal of Photogrammetry and Remote Sensing}, vol. 171, pp. 188--201, 2021.

\bibitem{dosovitskiy2020image}
A.~Dosovitskiy, ``An image is worth 16x16 words: Transformers for image recognition at scale,'' \emph{arXiv preprint arXiv:2010.11929}, 2020.

\end{thebibliography}

\end{document}